\documentclass[conference]{IEEEtran}
\ifCLASSINFOpdf
\else
\fi
\usepackage{fancyhdr} 
\usepackage{graphicx}
\usepackage{comment}
\usepackage{subfig}
\usepackage{algorithm}
\usepackage[noend]{algorithmic}
\usepackage{todonotes}
\usepackage{paralist}
\usepackage{wrapfig}
\usepackage{bbm}
\usepackage{xspace}
\usepackage{bbold}

\usepackage{newfloat}
\usepackage{listings}

\begin{document}

\bstctlcite{IEEEexample:BSTcontrol}
\title{Goal Space Abstraction in Hierarchical Reinforcement Learning via Reachability Analysis}


%%%%%%%%%%%%%%%  FILL OUUTTT %%%%%%%%%%%%%%%%%%%%%%%%%%%%%%%%%%%%%


\author{\IEEEauthorblockN{Mehdi Zadem\IEEEauthorrefmark{1}\IEEEauthorrefmark{2},
Sergio Mover\IEEEauthorrefmark{1},
Sao Mai Nguyen\IEEEauthorrefmark{2}}
\IEEEauthorblockA{\IEEEauthorrefmark{1}LIX, CNRS, École Polytechnique, Institute Polytechnique de Paris, Palaiseau, France}
\IEEEauthorblockA{\IEEEauthorrefmark{2}Flowers Team, U2IS, ENSTA Paris, Institut Polytechnique de Paris \& Inria, Palaiseau, France}
}


\maketitle
\thispagestyle{fancy}
\lhead{}
\chead{
\texttt{\scriptsize{Mehdi Zadem, Sergio Mover, Sao Mai Nguyen. Goal Space Abstraction in Hierarchical Reinforcement Learning via Reachability Analysis. Intrinsically Motivated Open-ended Learning IMOL 2023, Sep 2023, Paris, France. 
 }}
\vspace{20pt}}
\rhead{}
\cfoot{}
\begin{abstract}
Open-ended learning benefits immensely from the use of symbolic methods for goal representation as they offer ways to structure knowledge for efficient and transferable learning.
However, the existing Hierarchical Reinforcement Learning (HRL) approaches relying on symbolic reasoning are often limited as they require a manual goal representation.
The challenge in autonomously discovering a symbolic goal representation is that it must preserve critical information, such as the environment dynamics. 
In this work, we propose a developmental mechanism for subgoal discovery via an emergent representation that abstracts (i.e., groups together) sets of environment states that have similar roles in the task. We create a HRL algorithm that gradually learns this representation along with the policies and evaluate it on navigation tasks to show the learned representation is interpretable and results in data efficiency.
\end{abstract}


% \fi
\IEEEpeerreviewmaketitle

\section{Introduction}
Symbol emergence is key for developmental learning to tackle the curse of dimensionality and scale up to open-ended high-dimensional sensorimotor space, by allowing symbolic reasoning, compositionality, hierarchical organisation of the knowledge, etc. While symbol emergence has been recently investigated for the sensor data, action symbolization can lead to a repertoire of various movement patterns by bottom-up processes, which can be used by top-down processes such as composition to form an action sequence \cite{Nguyen2021KI}, planning and reasoning for more efficient learning, as reviewed in \cite{Taniguchi2018ITCDS}. Sensorimotor symbol emergence thus is key to scaling up primitive actions into complex actions for open-ended learning, using compositionality \cite{Manoury2019PICHI} and hierarchy \cite{Duminy2018PIICRC}.

Action hierarchies are the core idea of Hierarchical Reinforcement Learning (HRL) that decomposes a task into easier subtasks. In particular, in Feudal HRL \cite{feudal} a high-level agent selects subgoals that a low-level agent learns to achieve. The performance of Feudal HRL depends on the "hierarchical division of the available state space" \cite{feudal}, the representation of the goals that the high level agent uses to decompose a task.
%
%While such representation is critical in Feudal HRL, 
Yet, only few algorithms learn it automatically \cite{DBLP:journals/corr/VezhnevetsOSHJS17}, while others either use directly the state space~\cite{nachum2018dataefficient} or manually provide a representation~\cite{hdqn,DBLP:conf/nips/ZhangG0H020}.
%
% problem: 
In this research, we tackle the problem of learning automatically, while learning the policy,
a discrete interpretable goal representation from continuous observations
that expresses the task structure for data-efficiency.

We introduce a novel goal space representation and a feudal HRL algorithm, GARA (Goal Abstraction via Reachability Analysis), that develops such a representation while simultaneously learning a hierarchical policy from exploration data. The representation emerges through a developmental process, gradually gaining precision from a bottom-up manner, 
by leveraging data acquired from exploration. 
This discretisation of the environment is used to orient top-down process of the goal-directed exploration, that in turn helps improving policies and this representation.

\section{Formulation}
The goal space $\mathcal{G}$ is formulated as a partition of the state space $\mathcal{S}$ into $n$ disjoint sets of states $\mathcal{G} = \{G_0, \dots, G_n \}$ s.t $\bigcup_{G \in \mathcal{G}}{G} = \mathcal{S}$ and and $\forall G,G' \in \mathcal{G}$, $G \cap G' = \emptyset$ if $G \neq G'$. We define $R_k(G,G')$ as the set of states reached when starting from a state in $G$ and applying the low-level policy $\pi^{Low}(s \in G, G')$ targeting $G'$ for $k$ steps.
This goal space should satisfy the reachability property: $\forall ~ G, G' \in \mathcal{G}, R_k(G,G') \subseteq G'$ or $R_k(G,G') \cap G' = \emptyset$. Intuitively, this property expresses that each goal $G$ would group together states with a similar role in the task in terms of their ability to reach other goals. Inversely, if only some states in $G$ manage to reach the target $G'$ then $G$ contains states having different roles. This means that environment dynamics are not completely captured. In the following section we present GARA (Goal Abstraction via Reachability Analysis) that concurrently learns a hierarchical policy and the abstract goal space. 

\section{Methodology}
GARA is a Feudal HRL algorithm that learns two policies; a high-level policy $\pi^{High}: \mathcal{S} \rightarrow \mathcal{G}$ selects goals $G_i \sim \pi^{High}(s)$, and a low-level goal-conditioned policy $\pi^{Low}: \mathcal{S} \times \mathcal{G} \rightarrow \mathcal{A}$ that learns how to best achieve these goals by choosing actions in the action space $\mathcal{A}$ s.t $a_t \sim \pi^{Low}(s,G_i)$. $\pi^{High}$ is rewarded by the environment reward, while $\pi^{Low}$ is rewarded with respect to its ability to reach the selected goal.

\begin{figure}
    \centering
    \includegraphics[width=0.4\columnwidth]{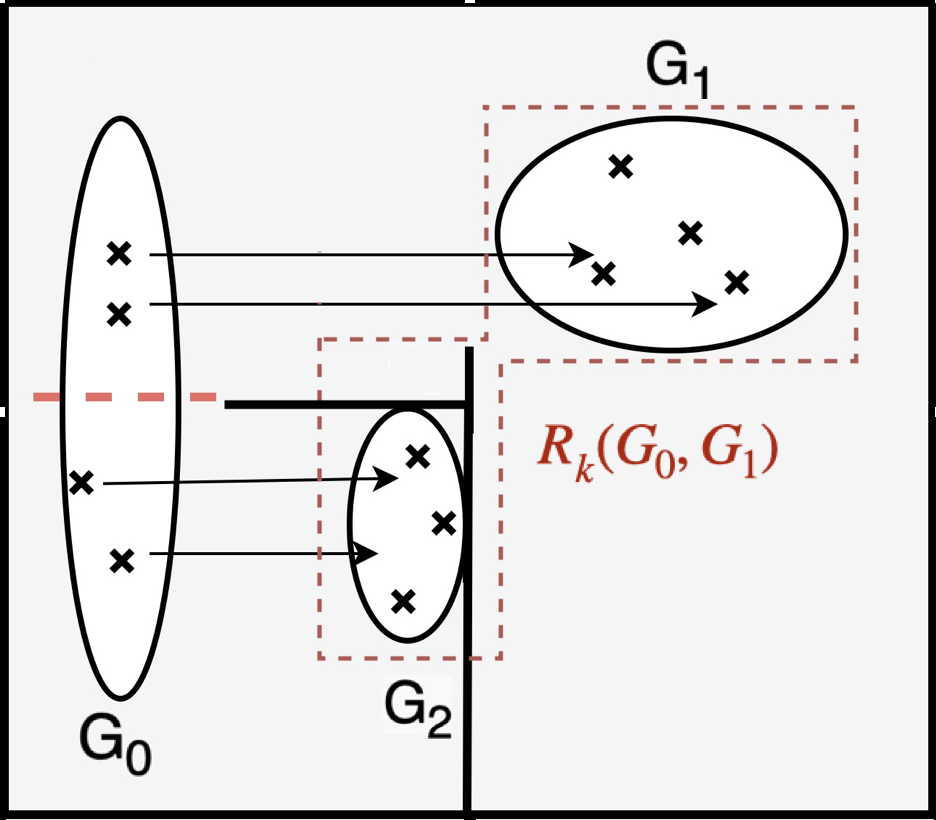}
    \vspace{-0.1cm}
    \caption{ In this maze, the transition starting from a state in $G_0$ with the policy "go right" may reach both $G_1$ and $G_2$. $G_0$ is thus split into two regions where all the states in the upper one reach $G_1$ and the states in the bottom one don't.}
    \label{fig:splitting}
    \vspace{-0.6cm}
\end{figure}

Learning the goal space comes down to identifying which states in each goal exhibit similar reachability behaviours. 
To this end, GARA trains a neural network from data acquired during exploration after each learning episode. This network is called the forward model $\mathcal{F}_k:\mathcal{S} \times \mathcal{G} \rightarrow \mathcal{S}$ such that $\mathcal{F}_k(s_t,G')$ predicts the state $s_{t+k}$ reached after applying $\pi^{Low}(s, G')$ for $k$ steps. A core idea of GARA, is that the reachability relations are computed over sets of states. To derive this from $\mathcal{F}_k$, we resort to a formal verification tool Ai2 \cite{DBLP:conf/sp/GehrMDTCV18} that can compute the output of a neural network given a set of inputs. More precisely, if the input to $\mathcal{F}_k$ is the set of states $G$, then the output should be an approximation of the reached set of states $\tilde R_k(G, G')$. For each explored transition from start set $G_s$ to destination goal $G_d$, Ai2 computes $\tilde R_k(G_s, G_d)$. If $\tilde R_k(G_s, G_d) \subseteq G_d$ or $\tilde R_k(G_s, G_d) \cap G_d = \emptyset$ then the reachability property is respected and $G_s$ would not be refined as the behaviour is similar across its states. Otherwise if $\tilde R_k(G_s, G_d) \not\subseteq G_d$ and $\tilde R_k(G_s, G_d) \cap G_d \neq \emptyset$, $G_s$ would be split in two sets $G'$ and $G''$ on which the reachability analysis is re-conducted (we compute $\tilde R_k(G', G_d)$ and $\tilde R(G'', G_d)$). This process continues recursively until the reachability relation is decidable. $G$ would thus be refined into two new sets $G'_s$ and $G''_s$ where $R_k(G'_s,G_d) \subseteq G_d$ and $\tilde R_k(G''_s, G_d) \cap G_d = \emptyset$. Fig.\ref{fig:splitting} illustrates this process.
The emerging regions constitute learning targets that are easily reachable and together would compose an abstract model for the task. 

\section{Results}
We focus on one experimental evaluation from our study which seeks to determine if an interpretable representation for the goal space can be learned from exploration, and if it helps the hierarchical policy to be more data-efficient. We conduct the evaluation on a U-shaped maze with a continuous state space, discrete actions controlling the agent's acceleration in 4 directions and a sparse reward is only attributed when reaching the exit. We compare GARA against some of the state-of-art approaches:

\noindent {\it - Feudal HRL with {\bf Handcrafted} representation:} inspired by hDQN \cite{hdqn}, this algorithm is similar in structure to GARA in using a discrete set-based goal space. This represntation is however handcrafted and fixed.

\noindent {\it - {\bf HIRO}:} also a feudal HRL algorithm, it relies on raw states to act as goals $\mathcal{G}=\mathcal{S}$. Additionally, it uses a goal interpolation mechanism along with hindsight experience.

\begin{figure}
    \centering
    \subfloat[Handcrafted representation \label{fig:U-shaped_handcrafted_repr}]{\includegraphics[width=0.2\columnwidth]{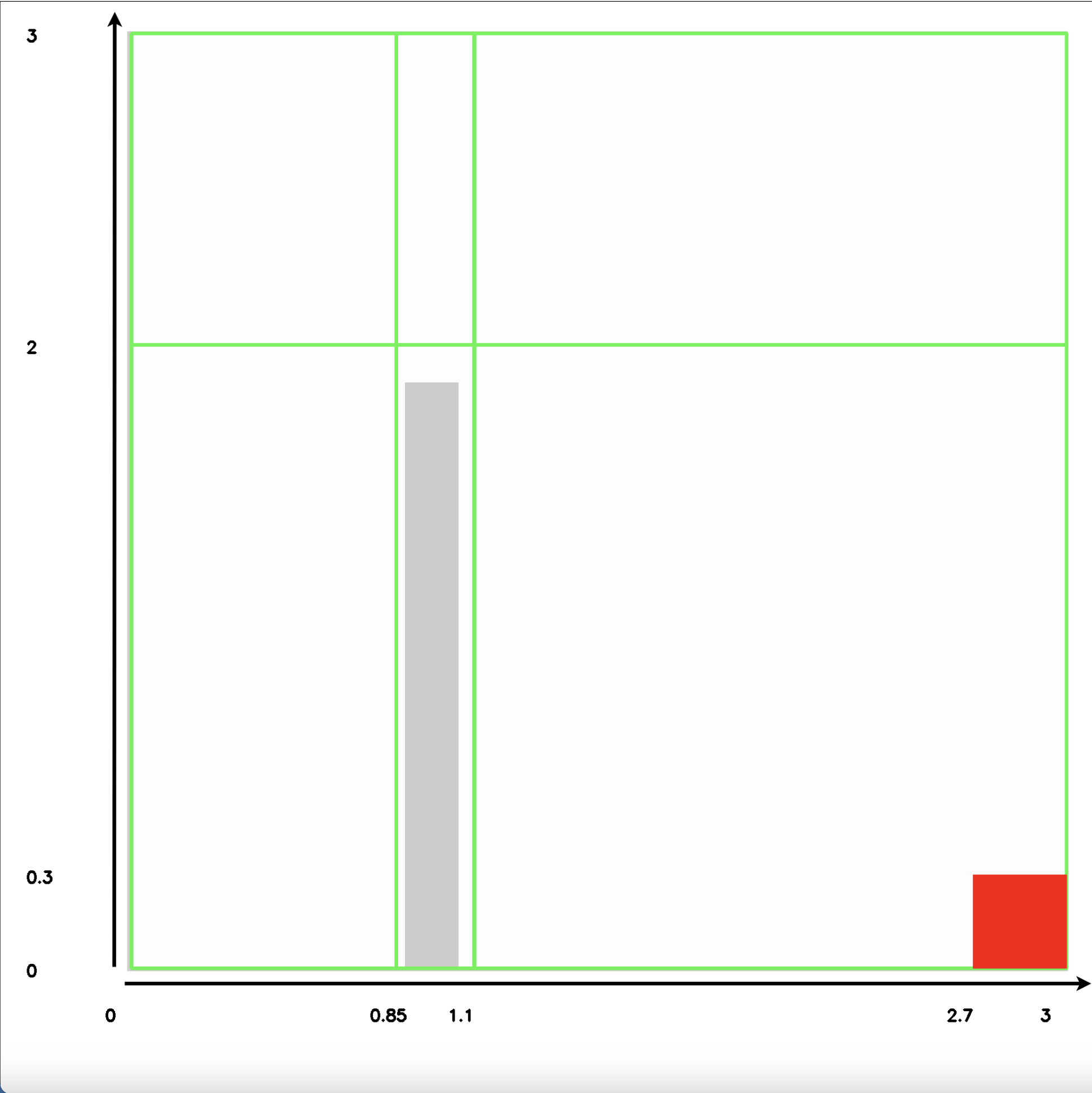}} 
    \hfill
    \subfloat[Initial $\mathcal{G}$ given to GARA \label{fig:U_shaped_repr1}]{\includegraphics[width=0.2\columnwidth]{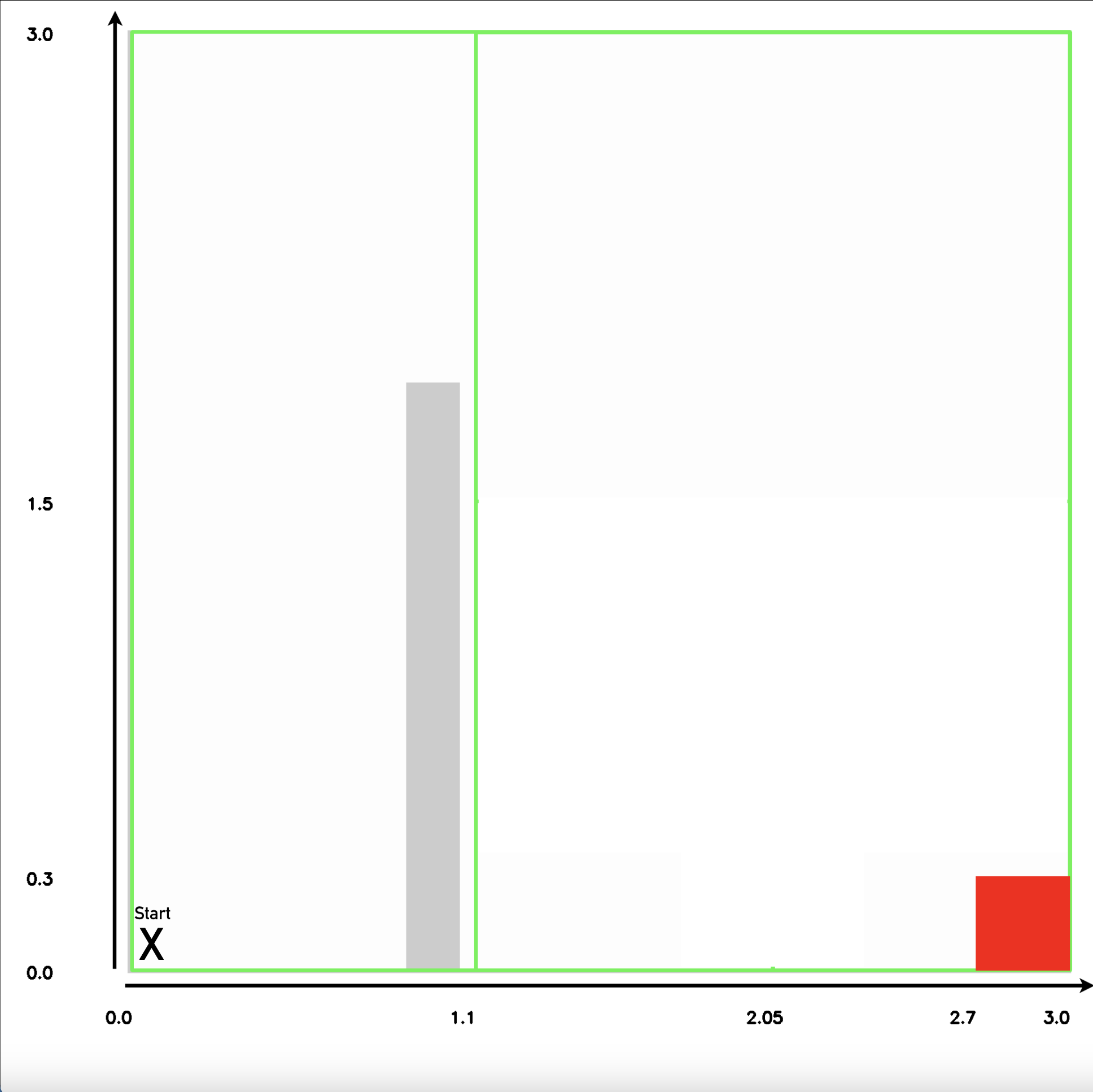}}
    \hfill
    \subfloat[$\mathcal{G}$ learned by GARA after $10^3$ steps \label{fig:U_shaped_repr2}]{\includegraphics[width=0.2\columnwidth]{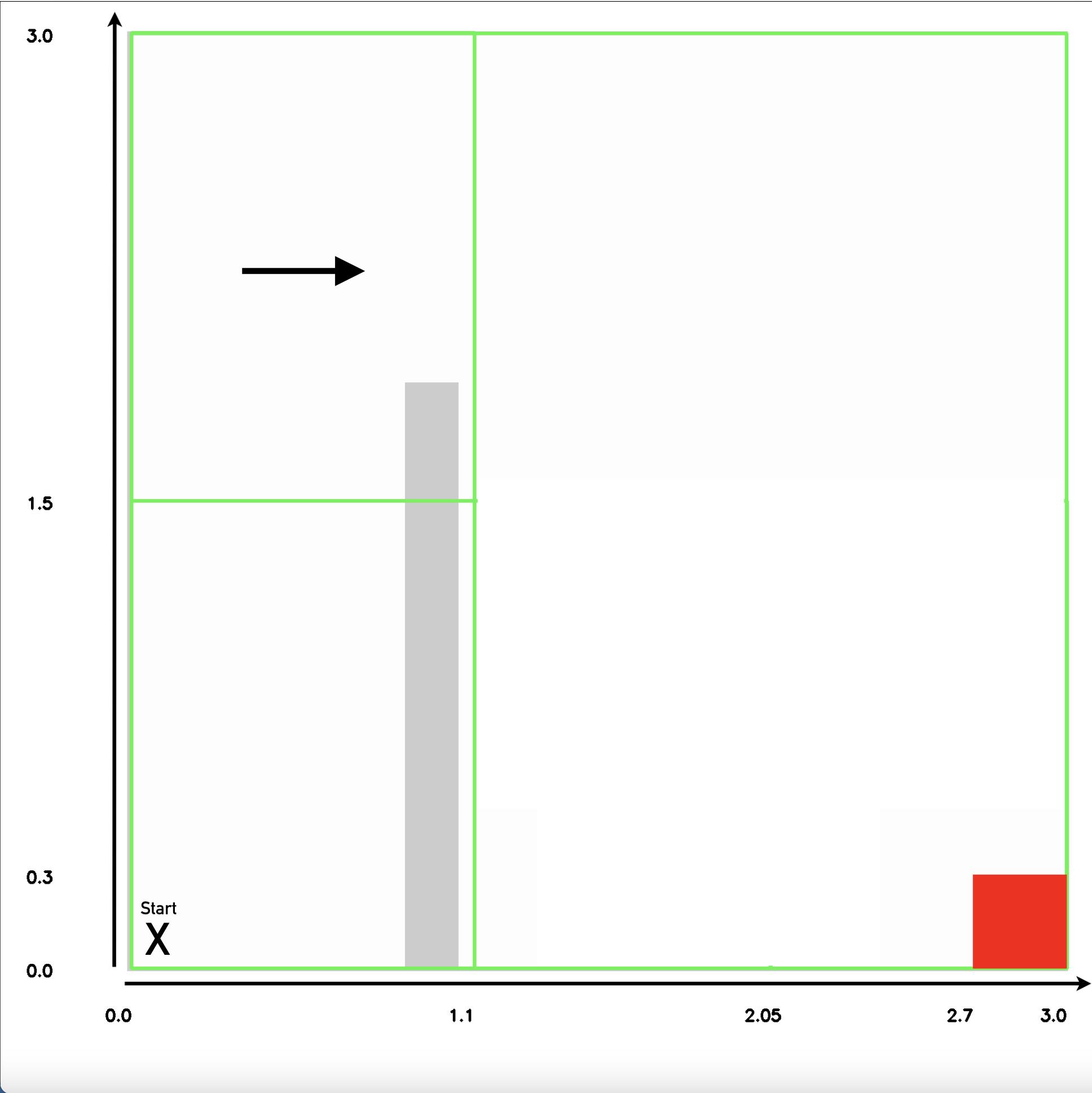}} 
    \hfill
    \subfloat[Final $\mathcal{G}$ learned by GARA \label{fig:U_shaped_repr3}]{\includegraphics[width=0.2\columnwidth]{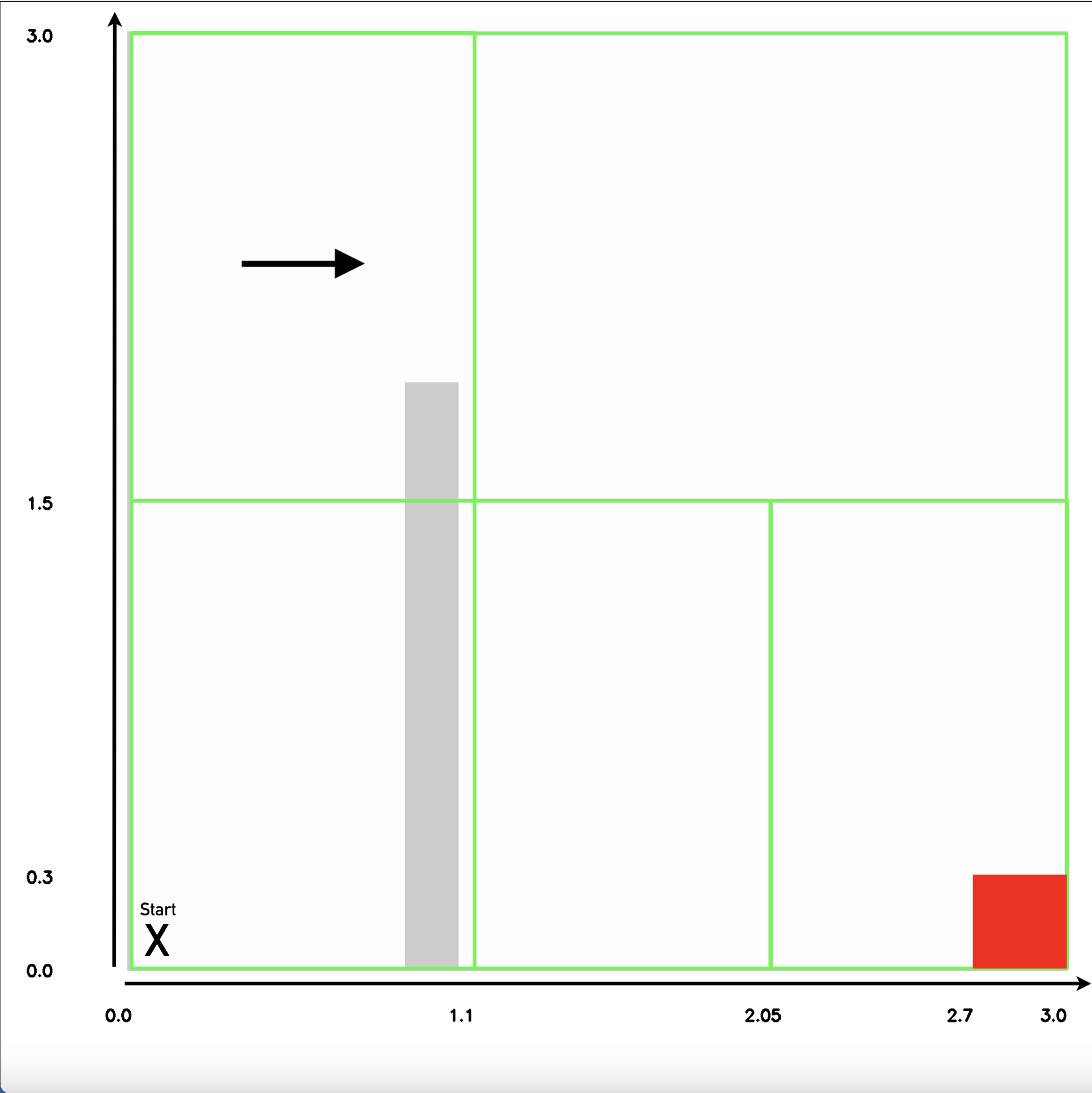}} 
    \vspace{-0.2cm}
     \caption{Representation of the goal space $\mathcal{G}$ in the U-shaped maze for one run of algorithm. The exit is marked in red.
     Green boxes show intervals for $x,y$ and the horizontal and vertical arrows indicate the sign of the velocities $v_x$ and $v_y$, respectively. No arrows indicate there are no split across $v_x$ or $v_y$.}
    \label{fig:U-shaped_repr}
    \vspace{-0.6cm}
\end{figure}

\paragraph{Representation learning} Focusing first on the learned representation by GARA, Fig.~\ref{fig:U_shaped_repr1}, Fig.~\ref{fig:U_shaped_repr2}, and Fig.~\ref{fig:U_shaped_repr3} show the evolution of the goal space throughout the learning at $0$, $10^3$, and $3 \times 10^4$ steps (for a randomly selected run of the algorithm).
Initially, GARA identifies the region at the top-left corner of the maze with positive velocity which provides a good starting point to learn policies that efficiently manage to reach the right half of the maze.
Later, GARA refines the right half region, which allows it to focus on the exit point.
Our intuition is that such final partition results in easier to reach goals, prompting the agent to select successful behaviours.
Overall, Fig.~\ref{fig:U-shaped_repr} shows that GARA learns an interpretable representation from data collected during the HRL exploration.

\begin{figure}
    \centering
    \includegraphics[width=0.8\columnwidth]{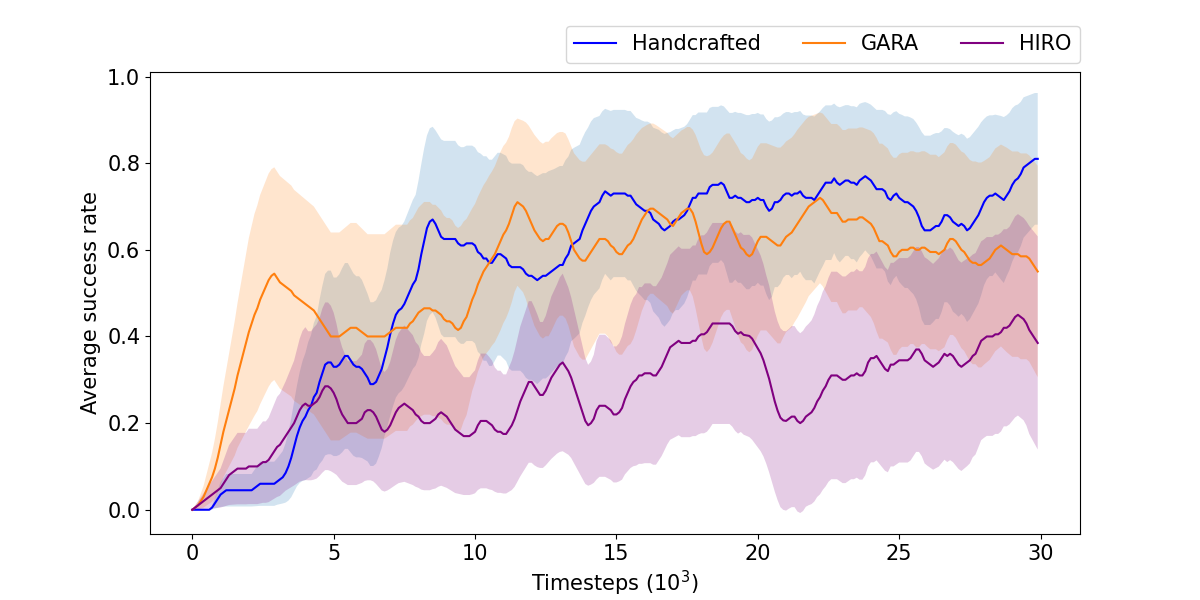}
    \vspace{-0.2cm}
    \caption{Average success rate on the U-shaped Maze (20 runs).}
    \label{fig:U_shaped}
    \vspace{-0.8cm}
\end{figure}

\paragraph{Data efficiency} Fig. \ref{fig:U_shaped} shows that our approach manages to learn a successful hierarchical policy with a performance approaching the handcrafted representation, whereas HIRO cannot learn to solve the task within the same time frame. We attribute this to the better sample efficiency associated with the learned abstraction, as the agents successfully decompose the task into simple-to-achieve goals.

\bibliographystyle{IEEEtran}
% (used to reserve space
%\bibliographystyle{plainnat}
\bibliography{references}

\begin{thebibliography}{10}
\providecommand{\url}[1]{#1}
\csname url@samestyle\endcsname
\providecommand{\newblock}{\relax}
\providecommand{\bibinfo}[2]{#2}
\providecommand{\BIBentrySTDinterwordspacing}{\spaceskip=0pt\relax}
\providecommand{\BIBentryALTinterwordstretchfactor}{4}
\providecommand{\BIBentryALTinterwordspacing}{\spaceskip=\fontdimen2\font plus
\BIBentryALTinterwordstretchfactor\fontdimen3\font minus
  \fontdimen4\font\relax}
\providecommand{\BIBforeignlanguage}[2]{{%
\expandafter\ifx\csname l@#1\endcsname\relax
\typeout{** WARNING: IEEEtran.bst: No hyphenation pattern has been}%
\typeout{** loaded for the language `#1'. Using the pattern for}%
\typeout{** the default language instead.}%
\else
\language=\csname l@#1\endcsname
\fi
#2}}
\providecommand{\BIBdecl}{\relax}
\BIBdecl

\bibitem{Nguyen2021KI}
S.~M. Nguyen, N.~Duminy, A.~Manoury, D.~Duhaut, and C.~Buche, ``Robots learn
  increasingly complex tasks with intrinsic motivation and automatic curriculum
  learning,'' \emph{K{\"u}nstliche Intelligenz}, vol.~35, 2021.

\bibitem{Taniguchi2018ITCDS}
T.~Taniguchi, E.~Ugur, M.~Hoffmann, L.~Jamone, T.~Nagai, B.~Rosman, T.~Matsuka,
  N.~Iwahashi, E.~Oztop, J.~Piater, and F.~Wörgötter, ``Symbol emergence in
  cognitive developmental systems: A survey,'' \emph{IEEE TCDS}, vol.~11,
  no.~4, pp. 494--516, 2019.

\bibitem{Manoury2019PICHI}
A.~Manoury, S.~M. Nguyen, and C.~Buche, ``Hierarchical affordance discovery
  using intrinsic motivation,'' in \emph{HAI}.\hskip 1em plus 0.5em minus
  0.4em\relax ACM, 2019.

\bibitem{Duminy2018PIICRC}
N.~Duminy, S.~M. Nguyen, and D.~Duhaut, ``Learning a set of interrelated tasks
  by using sequences of motor policies for a strategic intrinsically motivated
  learner,'' in \emph{Proceedings of IEEE International Conference on Robotic
  Computing}, 2018.

\bibitem{feudal}
P.~Dayan and G.~E. Hinton, ``Feudal reinforcement learning,'' in
  \emph{NeurIPS}, vol.~5, 1992.

\bibitem{DBLP:journals/corr/VezhnevetsOSHJS17}
A.~S. Vezhnevets, S.~Osindero, T.~Schaul, N.~Heess, M.~Jaderberg, D.~Silver,
  and K.~Kavukcuoglu, ``Feudal networks for hierarchical reinforcement
  learning,'' \emph{CoRR}, vol. abs/1703.01161, 2017.

\bibitem{nachum2018dataefficient}
O.~Nachum, S.~Gu, H.~Lee, and S.~Levine, ``Data-efficient hierarchical
  reinforcement learning,'' in \emph{NeurIPS 2018}, 2018.

\bibitem{hdqn}
T.~D. Kulkarni, K.~Narasimhan, A.~Saeedi, and J.~Tenenbaum, ``Hierarchical deep
  reinforcement learning: Integrating temporal abstraction and intrinsic
  motivation,'' in \emph{NeurIPS}, vol.~29, 2016.

\bibitem{DBLP:conf/nips/ZhangG0H020}
T.~Zhang, S.~Guo, T.~Tan, X.~Hu, and F.~Chen, ``Generating
  adjacency-constrained subgoals in hierarchical reinforcement learning,'' in
  \emph{NeurIPS}, 2020.

\bibitem{DBLP:conf/sp/GehrMDTCV18}
T.~Gehr, M.~Mirman, D.~Drachsler{-}Cohen, P.~Tsankov, S.~Chaudhuri, and M.~T.
  Vechev, ``{AI2:} safety and robustness certification of neural networks with
  abstract interpretation,'' in \emph{{IEEE} Symposium on Security and
  Privacy}, 2018.

\end{thebibliography}

\end{document}